\documentclass[letterpaper, 10 pt, conference]{ieeeconf}  

\IEEEoverridecommandlockouts                              

\usepackage{amsmath} 
\usepackage{amssymb}  
\usepackage[T1]{fontenc}
\usepackage{graphicx}
\usepackage{subfigure}
\usepackage{multirow}

\usepackage{booktabs}
\let\labelindent\relax
\usepackage{enumitem}

\title{\LARGE \bf
Fast and Accurate: An Adaptive VLA Inference Framework through Environment-aware Model Selection
}

\author{
    \authorblockN{
    Yuewei Sun$^{1,*}$, Lang Qin$^{1,*}$, Zechuan Tian$^{1}$, Jingwen Li$^{1}$, Guiqin Wang$^{1}$,\\
    Shengzeng Huo$^{1}$, Wenxin Ren$^{1}$, Tao Fang$^{1}$, Xiaochen Zhang$^{1}$, Guanqing Deng$^{1}$,\\
    Xiang Wang$^{1}$, Xiaowen Dong$^{1}$, Qinghai Guo$^{1}$, Yuxin Ma$^{2}$\\
    }
    \authorblockA{
    $^{1}$ACSLab, Huawei Technologies\\
    $^{2}$Southern University of Science and Technology\\
    {\tt\small qinlang24@huawei.com, guoqinghai@huawei.com, mayx@sustech.edu.cn}
}
\thanks{* These authors contributed equally to this work.}
}

\begin{document}

\maketitle
\thispagestyle{empty}
\pagestyle{empty}

\begin{abstract}

Embodied intelligence demands both long-horizon reasoning and real-time closed-loop responsiveness. Recent dual-system Vision–Language–Action (VLA) architectures combine fast reactive control with slow deliberative reasoning to balance inference speed and task success rate. However, existing dual-process VLAs tightly couple the fast module to intermediate representations of the slow module, necessitating end-to-end joint training and limiting modularity, extensibility and flexible system switching.
In this paper, we propose Environment-aware Model Selection (EMS), an adaptive VLA inference framework that switches between two fully decoupled systems of different scales through environment-aware model selection. The large-scale deliberative system provides globally consistent trajectory planning to ensure task success, while a lightweight reactive system enables high-frequency closed-loop control. A reinforcement-learning-based switching policy dynamically selects which system to invoke based on real-time feedback, enabling sparse use of the slow system and thereby balancing pretrained knowledge utilization with runtime efficiency.
Our design offers three key advantages over prior hierarchical VLA frameworks:
(1) a fully decoupled and modular dual-system architecture that supports plug-and-play model replacement;
(2) an adaptive, environment-aware switching strategy;
(3) high-frequency inference for responsive closed-loop control.
We extensively evaluate EMS in both simulation and real-world environments. On the LIBERO benchmark, EMS achieves success rates comparable to the large-scale baseline while increasing the effective action frequency to 93.4 Hz. The framework further demonstrates strong extensibility in real-world dual-arm manipulation tasks, where it accelerates task completion while maintaining robust performance.

\end{abstract}

\section{INTRODUCTION}

\begin{figure*}[t!]
    \vspace{5pt}
    \centering
    \includegraphics[width=1\linewidth]{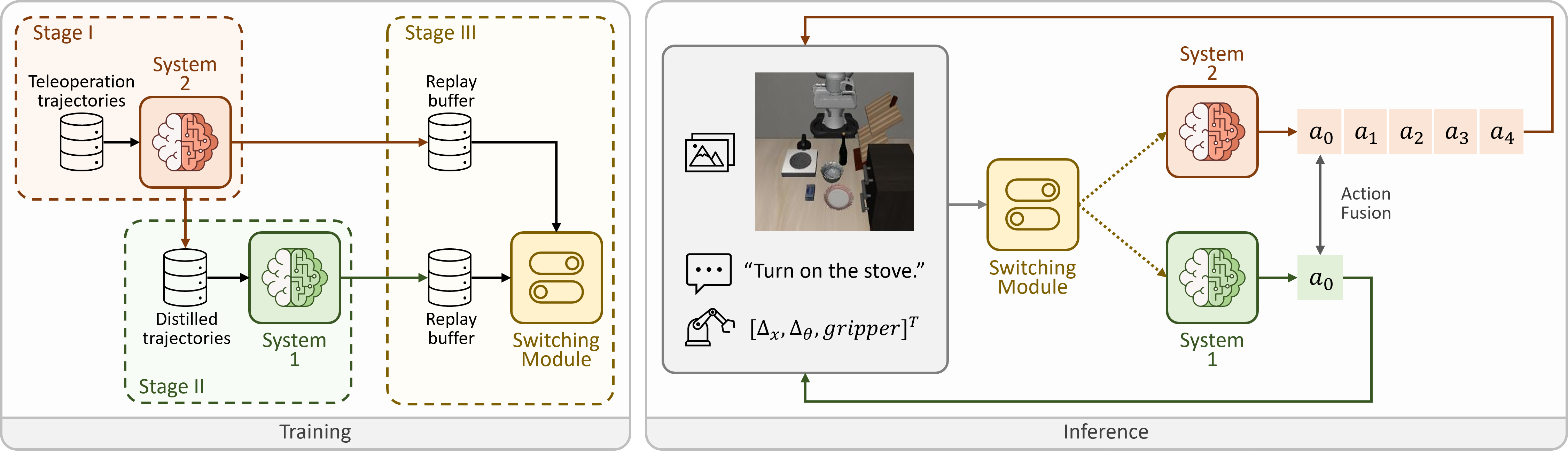}
    \vspace{-16pt}
    \caption{\textbf{Overview of the training and inference pipeline for the dual-system VLA framework.} Stage I trains System 2 on teleoperation trajectories, Stage II trains System 1 via trajectory distillation, and Stage III learns the switching module with reinforcement learning from replay buffers. During inference, the two systems operate independently, with the switching module dynamically selecting the appropriate system based on real-time observations. The action fusion process is employed to ensure smooth transitions between the two systems.}
    \label{fig:framework}
    \vspace{-12pt}
\end{figure*}

In recent years, vision–language–action (VLA) models\cite{pi0,openvla} have shown strong potential for vision-language-conditioned robotic control. 
With large parameter counts, they inherit rich priors from vision–language pretraining\cite{paligemma,prismatic} and further improve through continued training on large-scale robot datasets\cite{bridgedata,openX}.
As model scale increases, VLA models exhibit stronger spatial and task reasoning and improved generalization.
However, inference latency becomes a bottleneck under practical compute constraints, limiting control frequency and real-time closed-loop control.
For example, RT-2 (55B)\cite{rt-2} runs at 1–3 Hz, smaller VLA models like OpenVLA (7B)\cite{openvla} run at only 6.3 Hz, and diffusion-based policies\cite{diffusionpolicy,pi0} typically operate at around 10 Hz. 
These inference rates are often insufficient for responsive real-time control in dynamic, contact-rich manipulation. While action chunking can increase the effective action rate, it reduces the replanning frequency and delays reactions to unexpected feedback, since decisions remain bottlenecked by model inference in real deployments. This creates a fundamental trade-off between leveraging strong VLA performance and meeting stringent real-time execution requirements.

Addressing this bottleneck often calls for lightweight components. Directly shrinking VLAs is one option\cite{tinyvla,smolvla}, but it typically requires additional training and still struggles to match the latency of classical lightweight visuomotor policies such as BCTransformer\cite{BCTransformer} and ACT\cite{act}, which are reported to reach 100 Hz.
Inspired by Kahneman’s fast–slow theory\cite{thinking}, recent works adopt dual-system designs that combine a lightweight System 1 for high-frequency closed-loop control with a large, well-pretrained System 2 for deliberation and long-horizon reasoning.
Notable approaches such as DP-VLA\cite{dualvla} and FiS-VLA\cite{fast-in-slow} use System 2 to produce intermediate representations that are consumed by a co-trained System 1 for action prediction. While this can increase the effective execution rate, it introduces rigid cross-dependencies: the fast module cannot perform standalone end-to-end inference and must be jointly trained with the slow system, which hinders modular upgrades and precludes flexible, on-demand switching. Motivated by these limitations, we ask: {\bf{How can we build a dual-system VLA that (i) supports plug-and-play upgrades without expensive joint retraining, and (ii) achieves responsive real-time closed-loop control by selecting the appropriate level of computation based on the current environmental feedback?}}

To address this challenge, we propose EMS, an adaptive fast–slow VLA inference framework that balances execution speed and task success through environment-aware model selection.
As illustrated in Fig. \ref{fig:framework}, unlike prior dual-system approaches that rely on intermediate-state sharing, EMS restricts cross-system interaction to the action level, thereby avoiding feature-level dependencies and enabling each policy to be trained, deployed, and upgraded independently.
This model-agnostic interface allows plug-and-play replacement of both fast and slow policies and supports switching at arbitrary decision points during execution.
Specifically, System 1 is a lightweight reactive policy capable of end-to-end inference at approximately 100 Hz, whereas System 2 is a large pretrained deliberative policy that generates action chunks at around 10 Hz from a single observation.
To ensure extensibility across tasks and embodiments, both systems operate on the same observation space, including visual inputs and language instructions.
Leveraging this decoupled design, we further train a lightweight switching module that dynamically selects the active system online based on real-time environmental feedback.
This design enables efficient closed-loop control while retaining the planning capability of large-scale VLA models.

Maintaining trajectory consistency under fully decoupled operation and on-demand switching is non-trivial. We therefore introduce a stage-wise training pipeline. In Stage I, we fine-tune System 2 via imitation learning on task-specific trajectories collected through teleoperation or rule-based generation. In Stage II, we distill System 2 by training System 1 from scratch with imitation learning on successful rollouts produced by the fine-tuned System 2, aligning the fast policy with the slow policy’s trajectory distribution. In Stage III, we formulate system selection as a two-action Markov Decision Process (MDP) over robot state, where the controller selects either the fast or slow policy at decision boundaries.  We optimize the switching policy via reinforcement learning to maximize task success while discouraging unnecessary slowmodel invocation. To further stabilize execution at switching moments, we perform action fusion at the boundary by leveraging redundant actions from System 2’s planned chunk. With these designs, EMS achieves success rates close to System 2 while operating at an execution frequency comparable to System 1, requiring only sparse invocation of the slow system.
Our contributions are threefold:
\begin{itemize}[nosep, leftmargin=14pt, labelsep=7pt]
    \item \textbf{Decoupled dual-system inference.} We propose EMS, a plug-and-play dual-system VLA framework in which both the slow and fast modules support standalone end-to-end action generation and interact only at the action level, enabling modular upgrades without costly co-training.
    \item \textbf{Environment-aware adaptive switching.} We formulate system selection as a two-action decision problem and learn a lightweight switching policy with reinforcement learning to perform environment-aware model selection on demand.
    \item \textbf{Fast and accurate embodied control.} Extensive experiments in both simulation and real-world settings demonstrate that EMS achieves success rates comparable to large-scale VLA models while maintaining high execution frequency and reducing task completion time.
\end{itemize}

\section{Related Work}

\subsection{Vision-language-action models.}

Vision–language models (VLMs)\cite{clip} learn multimodal representations via large-scale pretraining on image–text corpora, aligning visual inputs with natural language. With the growth of robot datasets\cite{bridgedata,openX}, such pretrained models have been further adapted using large-scale robot data, enabling their semantic priors to transfer to embodied control\cite{r3m,vc-1}. Building on these robot-adapted VLMs, vision–language–action (VLA) models incorporate the action modality and are commonly trained with imitation learning on expert demonstrations\cite{cliport,BCTransformer}.
Early VLA models scaled up parameter counts and adopted autoregressive policies for discrete action prediction\cite{rt-2,openvla}. 
However, such methods may fail to ensure smooth action transitions and generally suffer from limited execution frequency.
To improve action continuity, several works incorporate diffusion-based action heads that generate temporally extended action chunks\cite{diffusionpolicy,pi0}, better capturing the multimodal structure of expert trajectories while producing smoother spatiotemporal behaviors. Nevertheless, diffusion-based policies typically still run at low inference frequencies, making it difficult to support responsive closed-loop control in dynamic environments.

\subsection{Dual-system VLAs.}
To improve the effective execution frequency of VLA models, recent approaches explore dual-system architectures. Following the fast–slow reasoning paradigm, System 2 typically denotes a large VLA model that provides deliberative decisions leveraging rich pretrained knowledge, while System 1 is a lightweight module intended for high-frequency closed-loop control. Some methods share intermediate representations between a large vision–language backbone and a small action head\cite{gr00t,pi0,cogact}; however, this tight feature-level coupling makes the end-to-end control rate a bottleneck of the large backbone, limiting the achievable speedup.
In addition, \cite{dualvla,hirt,robodual} adopt similar architectures but execute the modules asynchronously. By invoking the fast module at a higher rate, these methods seek to improve responsiveness and increase the effective control frequency, while relying on the slow module more sparsely for deliberation. Certain approaches further incorporate additional modalities\cite{fast-in-slow} or auxiliary modules\cite{trivla} and optimize them via joint training to achieve tighter inter-module coordination, at the cost of stronger coupling between components. Meanwhile, other methods improve task robustness by enabling the fast system to perform frequent replanning\cite{humevla}.
However, all of these approaches treat the fast system as a subordinate module that depends on slow-system features, leading to tight coupling between the two components. Such coupling limits modular upgrades and can bottleneck high-frequency control, reducing the benefits of asynchronous execution.
In this work, we propose EMS, an asynchronous dual-system framework in which both systems support independent end-to-end action inference and interact only through an action-level interface. A lightweight switching module selects which system to invoke based on real-time feedback, enabling each to operate according to its strengths.

\section{Method}

In this section, we present an overview of our proposed EMS framework, as shown in Fig. 1. In Section III-A, we introduce the problem formulation and solution approach. In Section III-B, we describe the overall system design, including the structure of System 1 ($\pi_{fast}$), System 2 ($\pi_{slow}$) and switching module, and explain how they coordinate during execution via environment-aware system selection. In Section III-C, we present the stage-wise training pipeline.

\subsection{Problem Formulation}

To combine deliberative planning with reactive execution, we adopt a high-capacity VLA model, a high-frequency VLA model, and an additional lightweight switching module.

Our goal is efficient and accurate robotic control, where both high success rates and fast execution are critical. In manipulation tasks, such as pick-and-place or object stacking, achieving both accuracy and speed is challenging due to environmental variability and the need for fine-grained closed-loop control. We focus on two key metrics:
\begin{itemize}[nosep, leftmargin=14pt, labelsep=7pt]
    \item {{\bf{Task success rate (S.R.)}}: Achieving reliable task completion, including in dynamic or unstructured environments.}
    \item {{\bf{Task Completion Time (T)}}: Minimizing task completion time in real-time closed-loop settings by enabling high-frequency feedback-driven control.}
\end{itemize}

To address these challenges, we propose a dual-system approach: a slow deliberative policy $\pi_{slow}$ for planning and a fast reactive policy $\pi_{fast}$ for rapid action execution. Both policies are standalone and support end-to-end action generation, i.e., given the current observation $o_t$, each can infer an executable action $a_t$ without relying on intermediate features from the other.
Both VLA models are trained with imitation learning, using demonstrations from different sources for the slow and fast systems.
However, simply having two independent policies is not sufficient, since robotic manipulation requires balancing execution speed and control precision depending on the current task state. To address this, we introduce a lightweight switching module that selects which policy to invoke at each decision boundary based on the current robot state $s_t$. A decision boundary occurs when the actions produced by the previous inference cycle (i.e., an action chunk or a single-step action) have been fully executed.

We formulate policy selection as an MDP, where the state is the robot state $s_t$, and the switching action is a discrete system selection action $\hat{a}_t \in \{0, 1\}$. Specifically, $\hat{a}_t = 0$ selects the fast system (System 1) with policy $\pi_{fast}$, and $\hat{a}_t = 1$ selects the slow system (System 2) with policy $\pi_{slow}$. After the selection, the chosen policy generates executable action(s) that are executed until the next decision boundary, resulting in the next state. 
The environment assigns an episodic reward $R_{total}$ where $R_{total}=r$ for success and $R_{total}=-r$ for failure. To mitigate reward sparsity, we distribute the total reward evenly across all time steps $T$, where the reward $r_t$ for time step $t$ is given by:
\begin{equation}
    r_t=\frac{R_{total}}{T}
\end{equation}
We optimize the switching policy $\pi(\hat{a}|s_t)$ using reinforcement learning.
The learned switching policy invokes the slow system sparingly—only when additional deliberation is beneficial—while defaulting to the fast policy for high-frequency reactive control. This enables EMS to achieve both a high task success rate and fast completion.

Finally, we distinguish between two types of actions: (1) {\bf{executable actions $a_t$}}, the continuous control commands output by $\pi_{fast}$ or $\pi_{slow}$. (2) {\bf{system selection actions $\hat{a}_t$}}, the discrete decisions produced by the switching module.

\subsection{Model Selection and System Execution Workflow}

{\bf{Model Selection.}}
To enable effective deliberative planning and high-frequency reactive execution, the two systems are designed to satisfy complementary functional requirements. Importantly, EMS is architecture-agnostic: System 1 and System 2 are specified by their functional roles rather than a fixed model design. Any models satisfying the below requirements can be integrated into our framework without feature-level coupling or cross-module co-training. This abstraction decouples system design from model selection, facilitating plug-and-play upgrades as VLA models evolve.

\begin{itemize}[nosep, leftmargin=14pt, labelsep=7pt]
    \item {\bf{Deliberative Slow System 2}}. 
    System 2 is expected to provide strong representation capacity and long-horizon reasoning for deliberative planning. In our framework, it is instantiated as a large, well-pretrained VLA model that leverages vision–language pretraining and generates temporally coherent action chunks {$a_t,a_{t+1},...,a_{t+c}$} for global trajectory guidance.
    System 2 prioritizes planning reliability and global task consistency, and is therefore designed to operate at a relatively low inference frequency.
    \item {\bf{Reactive Fast System 1.}} 
    System 1 is designed for low-latency control and high-frequency inference. Unlike tightly coupled dual-system architectures, our fast model supports independent end-to-end action generation, which enables flexible environment-aware switching. To facilitate seamless trajectory continuation, it shares the same observation and instruction as the slow model. While flexible in output format (single-step or short chunk prediction), the fast model is required to sustain high-frequency action generation—ideally near 100 Hz per step—to ensure responsive closed-loop control and efficient task completion.
    \item {\bf{Lightweight Switching Module}}.
    The switching module can be instantiated with different reinforcement learning formulations. To avoid introducing a computational bottleneck in high-frequency control, we implement it as shallow fully connected networks, ensuring negligible inference overhead compared to the policy networks. To keep the switching module lightweight and avoid redundant computation, we restrict its input to the robot state $s_t$, rather than the full multimodal observation $o_t$.
\end{itemize}

{\bf{Framework Execution Workflow.}}
The execution process follows a closed-loop control routine with three sequential steps.
\begin{itemize}[nosep, leftmargin=14pt, labelsep=7pt]
    \item {\bf{Step I: Switching Decision.}} At each decision boundary, the switching module observes the current state $s_t$ and selects which system to invoke for the next inference cycle.
    
    \item {\bf{Step II: Policy Inference.}} Given the selected system, the corresponding policy performs inference conditioned on the current observation $o_t$, producing either a single action or an action chunk.
    \item {\bf{Step III: Action Execution.}} The predicted action(s) are then executed, and the next switching decision is made only after the current output sequence (single action or chunk) has been fully executed, ensuring temporal consistency.
    
\end{itemize}
These steps repeat at decision boundaries throughout the episode: after executing the action (or action chunk), the system receives updated observation $o_{t'}$, and invokes the switching module again. The loop continues until task termination, enabling stable closed-loop control with on-demand system selection.

{\bf{Action Fusion.}}
At the switching time step $t$, mismatches between the two systems' action predictions can introduce control discontinuities. To stabilize execution at switching boundaries, we perform action fusion at the handoff step by averaging one action from each system. Formally, we fuse actions at the switching boundary by averaging the fast and slow actions:
\begin{equation}
\label{deqn_ex1a}
a_{t} =\frac{ a_{t}^{{fast}} + a_{t}^{{slow}}}{2}.
\end{equation}
When switching from the slow system to the fast system, we average the slow system’s overlapping action from its planned chunk with the fast system’s action inferred from the latest observation. When switching from the fast system to the slow system, the fast system performs one additional inference step.

\subsection{Stage-wise training pipeline}
{\bf{Stage I: Deliberative System 2 Finetuning}}.
Given that System 2 is responsible for maintaining global task consistency and ensuring high success rates, it must fully exploit the rich prior knowledge obtained from large-scale pretraining. To achieve this, we fine-tune the pretrained model via imitation learning on task-specific rollouts $\mathcal{D}_{tele}$ collected through teleoperation. The optimization objective is: 
\begin{equation}
\underset{\theta_{2}}{\max} \mathbb{E}_{(a_t,o_t,l) \sim \mathcal{D}_{tele}}[\log\pi_{slow}(a_t|o_t,l)]    
\end{equation}
This approach preserves the model’s pretrained representations while aligning its behavior with the target task distribution.

\begin{figure}[t!]
    \vspace{6.5pt}
    \centering
    \includegraphics[width=0.9\linewidth]{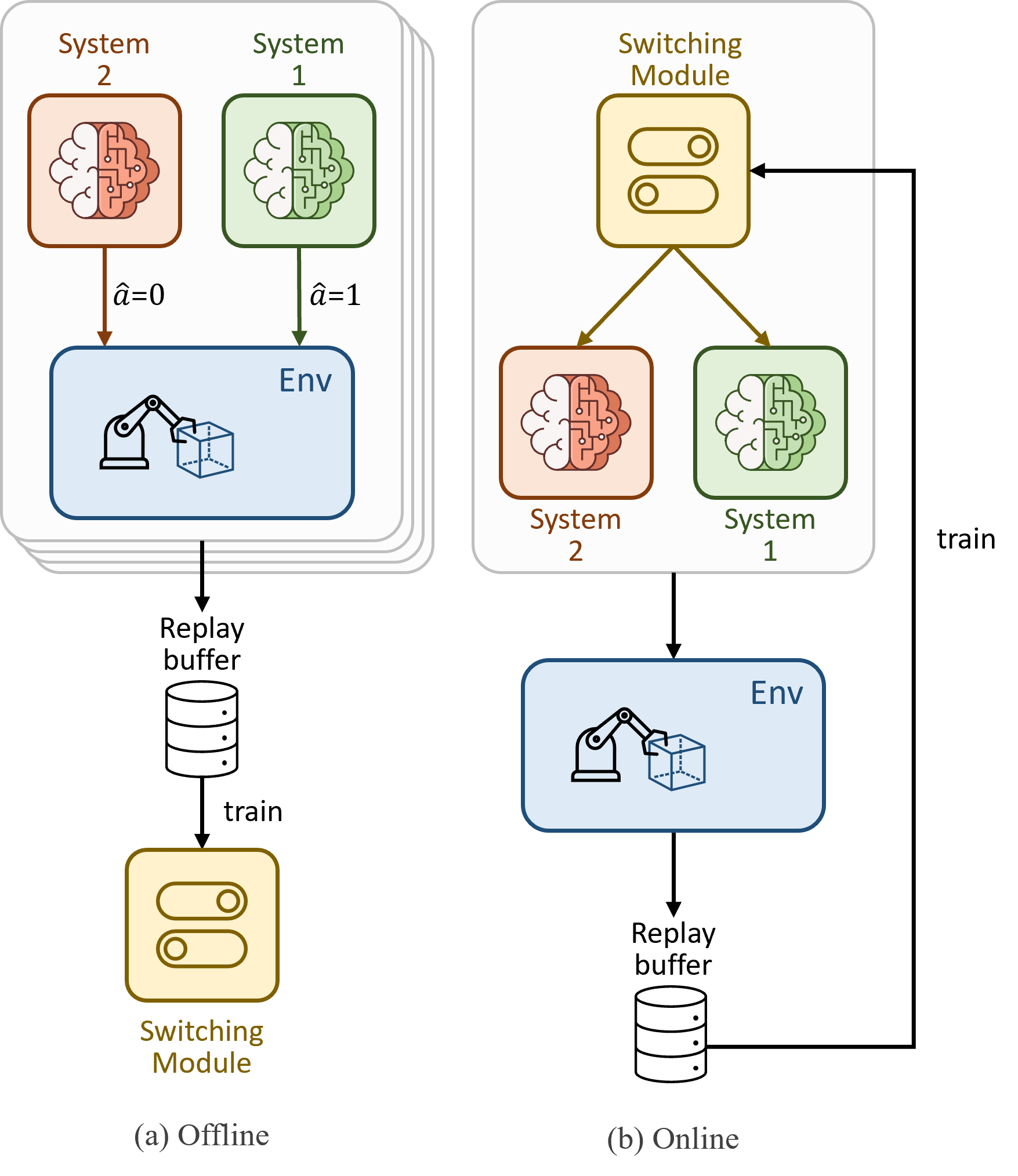}
    \vspace{-10pt}
    \caption{(a) \textbf{Offline training.} System 1 and System 2 independently interact with the environment to collect transition tuples, which are stored in a replay buffer. The switching module is trained using offline reinforcement learning based on these trajectories.
    (b) \textbf{Online training.} The switching module selects either System 1 or System 2 according to the current state, and the selected system interacts with the environment. The resulting transitions are used to further refine the switching policy, enabling adaptive system selection during execution.}
    \label{fig:reinforcement pipeline}
    \vspace{-12pt}
\end{figure}

{\bf{Stage II: Reactive System 1 Distillation}}.
System 1 is designed for high-frequency reactive execution and often continues trajectories initially planned by System 2. However, due to differences in model capacity and representation, independently trained models tend to learn distinct trajectory preferences under imitation learning. Such discrepancies can result in abrupt transitions at switching points, manifested as sudden acceleration or directional deviation, which adversely affects task stability and success rate.

Since System 2 primarily ensures global task success, it is essential that System 1 align with the trajectory distribution induced by System 2. To achieve this alignment, we adopt a trajectory distillation strategy. Specifically, the fine-tuned System 2 is used to generate rollouts $\mathcal{D}_{distill}$, and System 1 is trained from scratch using imitation learning on these distilled trajectories. The optimization objective is: 
\begin{equation}
\underset{\theta_{1}}{\max} \mathbb{E}_{(a_t,o_t,l) \sim \mathcal{D}_{distill}}[\log\pi_{fast}(a_t|o_t,l)]    
\end{equation}
By learning directly from the behavior distribution induced by System 2, System 1 minimizes trajectory mismatch and ensures smoother switching between the two systems.

{\bf{Stage III: Reinforcement Learning for Switching Module}}.
To address the challenge of real-time interaction between the model and the environment in real-world settings, we adopt both offline and online reinforcement learning strategies, enabling the system to adapt seamlessly to both simulation and real-world tasks. The online and offline reinforcement learning pipelines are illustrated in Fig. \ref{fig:reinforcement pipeline}.

\begin{table*}[t!]
    \vspace{6.5pt}
    \centering
    \caption{Success rate (S.R.), switch ratio ($\rho_{sw}$) and action frequency ($f_{action}$) on the LIBERO benchmark.}
    \vspace{-8pt}
    \begin{tabular}{c|c c c c|c|c|c}
        \toprule
        Models & LIBERO-Object & LIBERO-Spatial & LIBERO-Goal & LIBERO-10 & Mean S.R.(\%) &  Mean $\rho_{sw}$ &$f_{action}$(Hz)\\
        \midrule
        BCTransformer & 78.00 & 83.00 & 80.00 & 42.40 & 70.85 & - & 76\\
        BCVILT & 93.80 & 89.80 & 87.60 & 76.60 & 86.95 & - & 100\\
        OpenVLA & 84.70 & 88.40 & 79.20 & 53.70 & 76.50 & - & 6.3\\
        PI0 & \bf{98.80} & \bf{96.80} & \bf{95.80} & 85.20 & \bf{94.15} & - & 50\\
        \midrule
        Fixed Switching & 96.00 & 94.00 & 89.60 & 81.80 & 90.35 & 0.33 & 87.5\\
        EMS (ours) & 98.60 & 93.40 & 92.20 & \bf{85.40} & 92.40 & \bf{0.153} & \bf{93.37}\\
        \bottomrule
    \end{tabular}
    \vspace{-4pt}
    \label{tab:lib_results}
\end{table*}

\begin{figure*}[t!]
    \centering
    \includegraphics[width=1\linewidth]{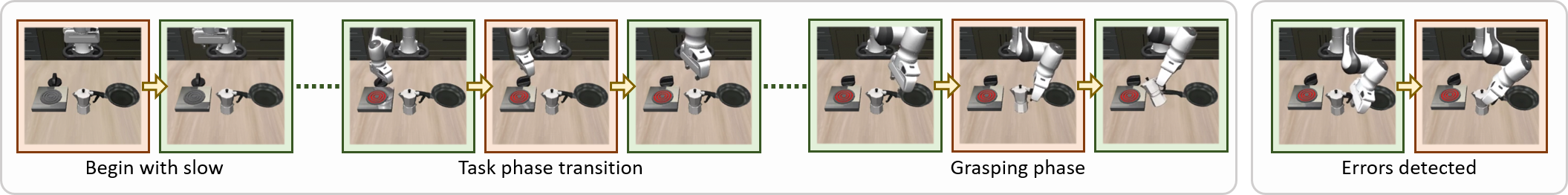}
    \vspace{-18pt}
    \caption{{\bf{Adaptive switching behavior in simulation.}}
    Green frames represent actions generated by the fast system, and orange frames represent actions generated by the slow system. The slow system is activated at critical stages (e.g., initial planning, grasping, phase transition, and error correction), while the fast system dominates routine trajectory execution.}
    \label{fig:switching behavior}
    \vspace{-14pt}
\end{figure*}


\textbf{Online Reinforcement Learning (DQN).}
To enable real-time switching under interactive settings, we adopt an enhanced Deep Q-Network (DQN) framework\cite{dqn}. Since the switching problem is formulated as a two-action discrete Markov Decision Process, DQN provides a computationally efficient solution suitable for high-frequency decision-making. To train the switching module online, we follow a standard collect–sample–update pipeline. During interaction with the environment, transitions ($s_t,\hat{a}_t,s_{t+1},r_t$) are stored in a replay buffer. To improve sample efficiency, we adopt prioritized experience replay, where each transition is sampled with probability proportional to its TD-error magnitude. For each sampled transition, the target value is computed using Double Q-learning to reduce overestimation bias in Q-value updates. The target value is computed as:
\begin{equation}
    y_t=r_t + \gamma Q_{\theta^{'}}(s_{t+1},\underset{\hat{a}}{argmax} Q_{\theta}(s_{t+1},\hat{a}))
\end{equation}
where $\theta$ and $\theta^{'}$ denote the online and target network parameters, respectively. To improve value estimation stability, the Q-network adopts a dueling architecture, decomposing the action-value function as
\begin{equation}
    Q(s_t,\hat{a}_t)=V(s_t)+(A(s_t,\hat{a}_t) - \frac{1}{|\mathcal{A}|}\sum_{\hat{a}'} A(s_t, \hat{a}'))
\end{equation}
The network is optimized by minimizing the squared TD-error,
\begin{equation}
    \mathcal{L}(\theta)=\mathbb{E}[(y_t - Q_{\theta}(s_t,\hat{a}_t))^2]
\end{equation}
For exploration, we employ NoisyNet to introduce learnable stochasticity in network parameters, enabling adaptive exploration without manual scheduling.
The target network is updated periodically to further stabilize training.

\textbf{Offline Reinforcement Learning (IQL).}
To enable offline switching policy learning, we adopt Implicit Q-Learning (IQL)\cite{iql}, which stabilizes learning by avoiding extrapolation error and performs implicit policy improvement through advantage-weighted regression. The state, action, and replay buffer settings are identical to those used in the online setup. IQL learns a Q-function and a state-value function without explicit policy constraint regularization. The $Q$-function update follows:
\begin{equation}
    \mathcal{L}_Q(\theta)=\mathbb{E}[(r_t+\gamma V_\psi(s_{t+1})-Q_\theta(s_t,\hat{a}_t))^2]
\end{equation}
and the value function $V(\psi)$ is optimized via expectile regression to mitigate overestimation:
\begin{equation}
    \mathcal{L}_V(\psi)=\mathbb{E}[\rho_\tau(Q_\theta(s_t,\hat{a}_t)-V_\psi(s_t))]
\end{equation}
where $\rho_\tau$ denotes the expectile regression loss. The policy is then implicitly derived by advantage-weighted behavioral cloning:
\begin{equation}
    \mathcal{L}_\pi(\phi)=\mathbb{E}[exp(\beta A(s_t,\hat{a}_t))log\pi_\phi(\hat{a}_t|s_t)]
\end{equation}
with $A(s_t,\hat{a}_t)=Q_\theta(s_t,\hat{a}_t)-V\psi(s_t)$.This design allows stable offline training without requiring explicit behavior policy constraints, making it well-suited for switching decisions learned from heterogeneous trajectories generated by both systems.

\section{Experiments}

We conduct comprehensive experiments in both simulation and real-world settings to evaluate the effectiveness, efficiency, and practicality of the proposed framework. In particular, we analyze task success rate,  action frequency and switching performance.
We define the switch ratio $\rho_{sw}$ as the ratio between the number of executable actions $N_{slow}$ produced by the slow policy and that $N_{fast}$ produced by the fast policy during an episode.
\begin{equation}
    \rho_{sw} = \frac{N_{slow}}{N_{fast}}
\end{equation}
The action frequency $f_{action}$ measures the effective rate of executable actions applied to the robot. For chunk-based policies, it is calculated as the chunk length $N_{chunk}$ multiplied by the model inference frequency $f_{infer}$.
\begin{equation}
    f_{action} = N_{chunk} \times f_{infer}
\end{equation}
Note that $f_{action}$ reflects the effective command rate under chunked execution and should not be interpreted as per-step closed-loop inference frequency. Moreover, in real-world systems the effective execution rate is jointly constrained by sensing, communication, and low-level control loops, making $f_{action}$ an unreliable proxy for end-to-end efficiency. Therefore, instead of reporting action frequency in the real world, we evaluate efficiency using the task completion time measured on the physical system.

\subsection{Simulation Experiment in LIBERO}

{\bf{Simulation benchmark.}}
To assess the effectiveness of our framework across diverse task settings, we evaluate it on four task suites in the LIBERO\cite{libero} simulation benchmark, including \textit{LIBERO-Object, LIBERO-Spatial, LIBERO-Goal} and \textit{LIBERO-10}. Each suite consists of 10 tasks specified by independent natural language instructions, and we report the suite-level performance as the average success rate across all tasks. All simulation experiments use a single Franka Panda robot. The observation $o_t$ is multimodal, including multi-view RGB images (front and wrist) as well as the end-effector pose and gripper state.

{\bf{Training and evaluation details.}}
In the single-arm setting, we adopt PI0\cite{pi0} as System 2 and BCVILT\cite{libero} as System 1 to learn a 7-dimensional action representation, where six dimensions correspond to the end-effector pose and one dimension controls the gripper state. System 2 produces an action chunk at each inference, whereas System 1 predicts a single action per time step.
Following the official PI0 implementation and its recommended setting on LIBERO, we set the System 2 (PI0) chunk size to $N_{chunk}=5$ throughout.

{\bf{In Stage I}} of training, PI0 is initialized from publicly available pretrained weights and fine-tuned on the official LIBERO teleoperation dataset, cleaned following the OpenVLA\cite{openvla} preprocessing pipeline.
We jointly train on four suites (40 tasks), with 50 successful trajectories per task (2,000 trajectories in total). Other training settings follow the PI0 released full fine-tuning configuration.
{\bf{In Stage II}}, we collect successful rollouts from PI0 on all 40 tasks using a fixed set of environment seeds, resulting in the same number of successful trajectories as in Stage I (50 per task; 2,000 in total). We then train BCVILT from scratch for 50 epochs on the collected trajectories.
{\bf{In Stage III}}, we train the switching module using online reinforcement learning.
We train a separate switching policy for each task suite.
Since the switching module is lightweight, its training overhead is negligible compared to training the VLA policies. For evaluation, we run 50 trials per task using held-out environment seeds that are disjoint from those used during training.

{\bf{Quantitative results.}}
As shown in Table \ref{tab:lib_results}, EMS achieves an average success rate of 92.40\% across four task suites, outperforming BCVILT by 5.45\%. 
We also evaluate a fixed switching baseline following the optimal setup of FiS-VLA, where the slow model is invoked once every four action chunks, yielding a switching ratio of 0.33.
Compared with the fixed-switching baseline, EMS reduces the average switching ratio from 0.33 to 0.153, while simultaneously increasing the success rate by 2.05\%. These results suggest that our adaptive selection invokes the slow model more sparingly, but does so at the right moments to improve task outcomes. Although EMS achieves a slightly lower average success rate than the standalone slow model overall, it marginally outperforms the slow model on the most challenging suite, LIBERO-10.

\begin{table*}[t!]
    \vspace{6.5pt}
    \centering
    \caption{Success rate (S.R.) and switch ratio($\rho_{sw}$) and task completion time across different task settings on the realman robot.}
    \vspace{-8pt}
    \label{tab:realman_results}
    \begin{tabular}{l|cc|cc|cc|ccc}
        \toprule
        \multirow{2}{*}{System} 
        & \multicolumn{2}{c|}{Single-arm simple} 
        & \multicolumn{2}{c|}{Single-arm hard}
        & \multicolumn{2}{c|}{Dual-arm simulation}
        & \multicolumn{3}{c}{Dual-arm real} \\
        & S.R.(\%) & $\rho_{sw}$ & S.R.(\%) & $\rho_{sw}$ & S.R.(\%) & $\rho_{sw}$ & S.R.(\%) & $\rho_{sw}$ & Task completion time(s) \\
        \midrule
        System 1 & 74 & - & 20 & - & 80 & - & 60 & - & 18 \\
        System 2 & 83 & - & \textbf{56} & - & 93 & - & \textbf{100} & - & 29 \\
        EMS      & \textbf{84} & 0.34 & 50 & 1.68 & \textbf{94} & 1.48 & 70 & 1.0 & 23 \\
        \bottomrule
    \end{tabular}
    \vspace{-10pt}
\end{table*}

{\bf{Effectiveness of Adaptive Switching.}}
We further analyze the switching behavior of the adaptive policy. The results show clear preferences in when the slow system is invoked, suggesting that the switching module captures task-relevant signals from the robot state.
As shown in Fig. \ref{fig:switching behavior}, the policy typically invokes the slow system at the beginning of an episode to obtain a reliable initial plan and avoid large early deviations, and then relies on the fast system for the majority of execution. Switching to the slow system is most commonly observed in three scenarios: 
\begin{itemize}[nosep, leftmargin=14pt, labelsep=7pt]
    \item During critical fine manipulation tasks, such as the grasping phase in pick-and-place, where precise decisions are crucial for task success.
    \item At task phase transitions, for example, in LIBERO-10, after completing one sub-goal, the policy invokes the slow system to plan the next.
    \item When errors are detected in action execution, prompting the system to switch back for replanning and correction.
\end{itemize}
These patterns indicate that the switching strategy is task- and state-aware rather than adhering to a fixed schedule.

\subsection{Ablation Study}

\begin{figure}[t!]
    \centering
    \includegraphics[width=0.9\linewidth]{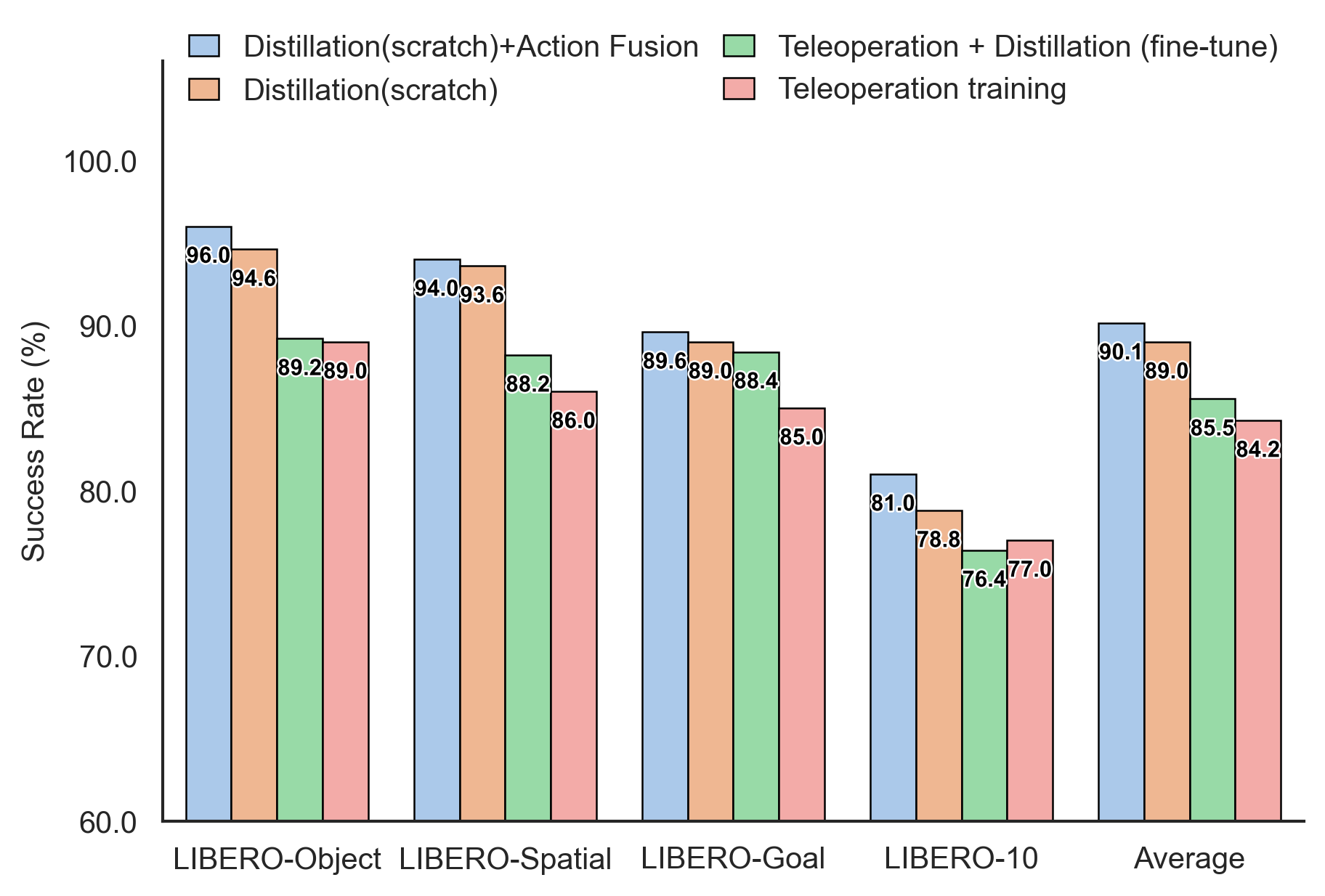}
    \vspace{-8pt}
    \caption{{\bf{Ablation study on System 1 training strategies and action fusion.}}
    Training System 1 from scratch on distillation trajectories performs best; teleoperation-only or teleoperation-then-distill underperforms, and removing action fusion consistently degrades success, especially on long-horizon tasks.}
    \label{fig:ablation result}
    \vspace{-14pt}
\end{figure}

To assess the impact of each component in the pipeline, we conduct ablation studies on LIBERO under the same setting as our simulation experiments. To ensure fair comparisons and isolate the effect of each component, we adopt a fixed switching schedule in all ablations. The quantitative results are summarized in Fig. \ref{fig:ablation result}.

{\bf{Effect of Trajectory Distillation.}}
Training the fast policy (System 1) on PI0-distilled trajectories consistently outperforms training on the original LIBERO teleoperation dataset. Using teleoperated data instead of distilled data leads to more than a 1\% performance drop on each of the four LIBERO suites and a 4.80\% reduction in the overall average success rate.
These results suggest that trajectory distillation is important for improving consistency between the fast and slow behaviors, thereby reducing switch-related trajectory inconsistencies and improving task success.

{\bf{Initialization Strategy for System 1.}}
We further fine-tune a teleoperation-trained System 1 on distilled trajectories, but observe only marginal gains. This suggests that the behavior learned from the initial training distribution largely determines the resulting policy, and subsequent fine-tuning on distilled data is insufficient to fully shift it toward the distilled trajectory distribution.

{\bf{Effect of Action Fusion.}}
Without action fusion at the switching handoff, the average success rate drops by 1.1\%, with a 3\% reduction on the LIBERO-10 suite. These results suggest that action fusion helps maintain smoother trajectories across switching boundaries, with a larger benefit on long-horizon tasks.

\subsection{Simulation Experiment in Realman}

\begin{figure}[t!]
    \centering
    \includegraphics[width=1\linewidth]{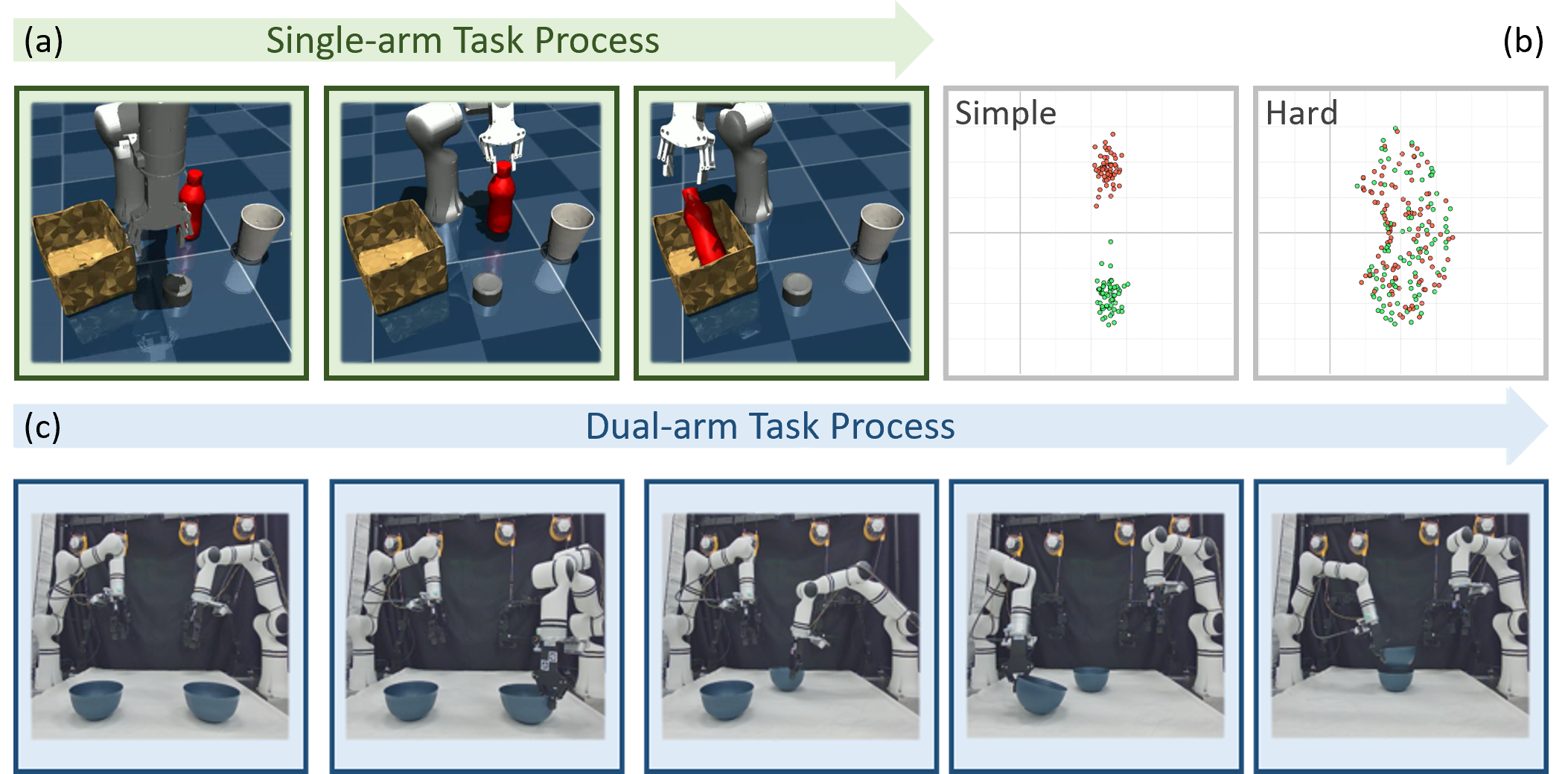}
    \vspace{-14pt}
    \caption{
    (a) \textbf{Single-arm task process.} Including grasping and placement, executed by the single-arm robot in temporal order. (b) \textbf{Initial object configurations.} Showing different initial placements of the bottle and the basket under the simple and hard settings. (c) \textbf{Dual-arm task process.} Including grasping, placement, and stacking, executed by the dual-arm system in temporal order.}
    \label{fig:realman task setting}
    \vspace{-14pt}
\end{figure}

To evaluate the portability of our framework across different robot embodiments and to support real-robot experiments, we conduct tests in a MuJoCo simulation environment using a model of the Realman RM75-6F manipulator. The evaluation includes both single-arm and dual-arm settings. We keep System 2 unchanged and use PI0 throughout. For System 1, we use BCVILT for single-arm tasks, and adopt ACT for dual-arm tasks to better support bimanual control. In the simulation environment, we continue to train the switching module with online reinforcement learning. Task processes are shown in Fig. \ref{fig:realman task setting}.

{\bf{Single-arm.}} For the single-arm setting, we consider a basic pick-and-place task that requires grasping a bottle and placing it into a basket, with two difficulty levels. In the easy setting, similar to the LIBERO setup, both the bottle and the basket are randomized within a small region. In the hard setting, their initial positions are fully randomized within the robot’s reachable workspace. 
As shown in Table \ref{tab:realman_results}, on the easier tasks, EMS achieves a higher success rate than System 2 with an average switch ratio of 0.34, suggesting that high-frequency closed-loop execution and sparse deliberative intervention are complementary.
On the harder tasks, despite a large performance gap between System 1 and System 2, EMS still achieves a success rate close to System 2. However, it reaches this performance by invoking the slow system more often, reflecting the necessary trade-off under challenging settings.
Nevertheless, since the switch ratio reflects the fraction of slow-policy invocations, EMS still attains substantial speedups by executing a large portion of actions with the fast policy.

{\bf{Dual-arm.}} In the dual-arm setting, the simulation task mirrors the real-robot task described later, and EMS similarly outperforms the System 2 baseline in success rate. The learned switching policy similarly tends to invoke the slow system more frequently to ensure task success.

\subsection{Real-World Experiment}

{\bf{Robot Setup and Task.}} In our real-world experiments, we use two Realman RM75-6F manipulators to form a dual-arm system for testing the pick-and-place task. We follow the task setup from LIBERO to design a multi-stage task {\textit{Stack Bowls Two}}, which involves evaluating grasp point localization, precise placement, and long-horizon coordination. The task consists of four sequential stages: (1) the left arm grasps a plastic bowl, (2) places it at a fixed location, (3) the right arm grasps a second bowl and (4) stacks it on top of the first. The initial positions of the bowls are randomized within a 10 cm radius, thereby introducing controlled variability into task execution. The task is considered successful when both bowls are stably stacked, and both arms return to their initial positions.

{\bf{Training and evaluation details.}} 
In the dual-arm setting, System 2 remains PI0, while System 1 is replaced with ACT to better accommodate robot setup. Both models operate with a 20-step action chunk. Following the same training protocol described earlier, each model collects 100 successful trajectories for policy training. For training the switching module, we additionally collect 50 trajectories from each model independently for offline reinforcement learning. We conduct 10 test runs per model and report the corresponding success rate.

{\bf{Quantitative results}}
System 2 (PI0) achieves a 100\% success rate, while System 1 (ACT) reaches 60\%. In comparison, EMS achieves a success rate of 70\%. In real-world dual-arm scenarios, the performance of the system is more constrained by the fast system, with failures primarily occurring when the right arm, using the fast system, experiences significant deviations.
Unlike in the simulation environment, where actions are performed at fixed time intervals, real-world execution is influenced by real-time dynamics and system constraints. We observe a significant speed-up in task completion with our dual-system framework. Under the same experimental setup, System 2 independently completes the task in an average of 29 seconds, while EMS reduces this time to 23 seconds. Although the System 1 alone theoretically completes the task faster in 18 seconds, it often requires replanning due to lower trajectory quality, resulting in execution times of 30 seconds or more.

\section{Conclusion}

In this work, we introduce EMS, a decoupled dual-system VLA framework consisting of an independent slow deliberative model (System 2), a fast reactive model (System 1), and a lightweight switching module for environment-aware system selection. We develop a stage-wise training strategy that enables effective coordination and stable interaction between the two systems. By learning an adaptive switching policy, our framework achieves high-frequency control while maintaining strong task success rates. 
Extensive experiments in both simulation and real-world settings demonstrate that EMS balances accuracy and efficiency.


\bibliographystyle{IEEEtran}
\bibliography{IEEEabrv,IEEEexample}

\end{document}